\documentclass[english]{article}
\usepackage[latin9]{inputenc}
\usepackage{babel}
\usepackage{amsmath}
\usepackage{graphicx}
\usepackage[unicode=true,pdfusetitle,
 bookmarks=false,
 breaklinks=false,pdfborder={0 0 1},backref=false,colorlinks=false]
 {hyperref}

\makeatletter

\providecommand{\tabularnewline}{\\}

\usepackage[final,nonatbib]{nips_2016}
\usepackage{mathptmx}
\usepackage{capt-of}

\makeatother

\begin{document}

\title{Control Matching via Discharge Code Sequences}

\author{Dang Nguyen, Wei Luo, Dinh Phung, Svetha Venkatesh\\
Centre for Pattern Recognition and Data Analytics\\
School of Information Technology, Deakin University, Geelong, Australia\\
\{d.nguyen, wei.luo, dinh.phung, svetha.venkatesh\}@deakin.edu.au}
\maketitle
\begin{abstract}
In this paper, we consider the patient similarity matching problem
over a cancer cohort of more than 220,000 patients. Our approach first
leverages on Word2Vec framework to embed ICD codes into vector-valued
representation. We then propose a sequential algorithm for case-control
matching on this representation space of diagnosis codes. The novel
practice of applying the sequential matching on the vector representation
lifted the matching accuracy measured through multiple clinical outcomes.
We reported the results on a large-scale dataset to demonstrate the
effectiveness of our method. For such a large dataset where most clinical
information has been codified, the new method is particularly relevant.\vspace{-0.2cm}
\end{abstract}

\section{Introduction\vspace{-0.2cm}
}

Recently, using ICD codes \cite{icd10} encoded in Electronic Medical
Records (EMR) for patient similarity matching has attracted a lot
of attention \cite{carnaby2010mcneill,gottlieb2013method,lee2015personalized}.
The basic idea is that a patient (called a \textit{case}) is typically
paired with a clinically similar patient (called a \textit{control})
with respect to ICD code sequence. To determine the similarity of
two ICD code sequences, one can compare their primary diagnosis (i.e.,
their first ICD code) \cite{carnaby2010mcneill,lee2015personalized}
or compute their Hamming distance \cite{hielscher2014using}. A major
difficulty in comparing ICD code sequences is the variation in encoding,
i.e., different ICD codes can be used to record the same disease.
For example, both ICD codes I20.0 and R57.0 are related to \textquotedblleft heart
issue\textquotedblright . Another difficulty is the sequential importance
in the code sequences. Most existing approaches miss to leverage either
the clinical relation among syntactically different ICD codes or the
relative position of ICD codes within a sequence \cite{hielscher2014using,lee2015personalized}.

To incorporate these two sets of important information in patient
similarity matching, we here introduce WVM, a method that matches
sequences of ICD codes within the embedded vector space. The method
embeds ICD codes using Word2Vec \cite{mikolov2013distributed} to
capture semantic similarity among syntactically different codes. It
also has a new sequential matching technique that leverages the domain
coding convention to capture similarity among patients with complex
syndromes and comorbidities. Our proposed method not only addresses
the unavoidable coding variation problem but also considers the important
sequential structure in ICD code sequences. We evaluate the performance
of our model based on the similarity of the matched patients on two
sets of outcomes: 28-day readmission and death due to cancer. A good
matching should produce patient pairs similar in these two outcomes.
Our cohort consists of more than 220,000 patients whose data were
collected from a state-wide cancer registry in Australia\textemdash The
study constitutes the first step for a set of comparative observational
studies. While sharing a similarity with medical concept embedding
as in Code sum based matching (CSM) \cite{choi2016medical}, our framework
with the proposed sequential matching yields better matching result.
In comparison with recent non-embedding approaches such as Primary
code based matching (PCM) \cite{lee2015personalized} and Hamming
distance based matching (HDM) \cite{hielscher2014using}, we also
achieve a better performance.\vspace{-0.2cm}

\section{Word2Vec based matching (WVM)\vspace{-0.2cm}
}

The proposed WVM method for patient similarity matching has two phases.\vspace{-0.2cm}

\subsection{Phase 1: Learning ICD code vectors\vspace{-0.2cm}
}

WVM uses Word2Vec (Skip-gram model \cite{mikolov2013distributed})
to learn ICD code vectors that can capture the latent relations among
ICD codes. This idea was also used in \cite{choi2016learning,choi2016medical}.
The local context is defined to be ICD codes that appear in the same
episode (admission) with a window size of 5. Those ICD code vectors
are then used in a sequential matching algorithm in phase 2.\vspace{-0.2cm}

\subsection{Phase 2: Building a matching algorithm\vspace{-0.2cm}
}

The inputs of the sequential matching (SM) algorithm are a group of
cases (\textit{case group}) and a group of controls (\textit{control
group}). We will describe how we select case group and control group
in Section \ref{subsec:Construction-of-case}. SM has three main steps.
\textbf{Step 1:} For a given case $tc^{\left(i\right)}$ it finds
a \textit{validation group} that consists of the controls that have
the same gender and age group with $tc^{\left(i\right)}$. Matching
on factors such as gender and age is commonly used in case-control
studies \cite{pearce2016analysis}. \textbf{Step 2:} It generates
two sets $S$ and $matched$ to store the controls that match $t$
ICD codes in $tc^{\left(i\right)}$ ($t\in\left[0,n\right]$, where
$n$ is the number of ICD codes in $tc^{\left(i\right)}$). \textbf{Step
3:} It matches $tc^{\left(i\right)}$ with a control based on $S$
and $matched$. 

In steps 2 and 3, to find a matching control for $tc^{\left(i\right)}$,
there are three scenarios to consider.

\textbf{Scenario 1:} We can find the controls that match all ICD codes
in $tc^{\left(i\right)}$. SM selects randomly a control in $S$ for
it to be the matching control $mc$.

\textbf{Scenario 2:} We cannot find any control that matches all ICD
codes in $tc^{\left(i\right)}$ but we are still able to find the
controls that match at least one ICD code in $tc^{\left(i\right)}$.
Thus, $S$ is empty but $matched$ is not. Assume that $matched$
contains the controls that match $\left(k-1\right)$ ICD codes in
$tc^{\left(i\right)}$. Since we cannot find any control that matches
the $k^{th}$ ICD code in $tc^{\left(i\right)}$ (called $tc_{k}^{\left(i\right)}$),
SM tries to search for a control $vc^{\left(j\right)}$ whose the
$k^{th}$ ICD code (called $vc_{k}^{\left(j\right)}$) is similar
to $tc_{k}^{\left(i\right)}$. SM first obtains the ICD code vector
$v_{i}$ of $tc_{k}^{\left(i\right)}$. For each $vc^{\left(j\right)}$
in $matched$, it obtains the ICD code vector $v_{j}$ of $vc_{k}^{\left(j\right)}$.
It then computes the cosine distance between $v_{i}$ and $v_{j}$.
The matching control for $tc^{\left(i\right)}$ is the one with the
smallest distance. 

\textbf{Scenario 3:} We cannot find any control that matches the first
ICD code in $tc^{\left(i\right)}$. Thus, both $S$ and $matched$
are empty. SM attempts to match $tc^{\left(i\right)}$ with a control
in \textit{validation group} whose the first ICD code is similar to
the first ICD code in $tc^{\left(i\right)}$ by computing the distance
between them.

\begin{figure}[th]
\begin{centering}
\includegraphics[scale=0.4]{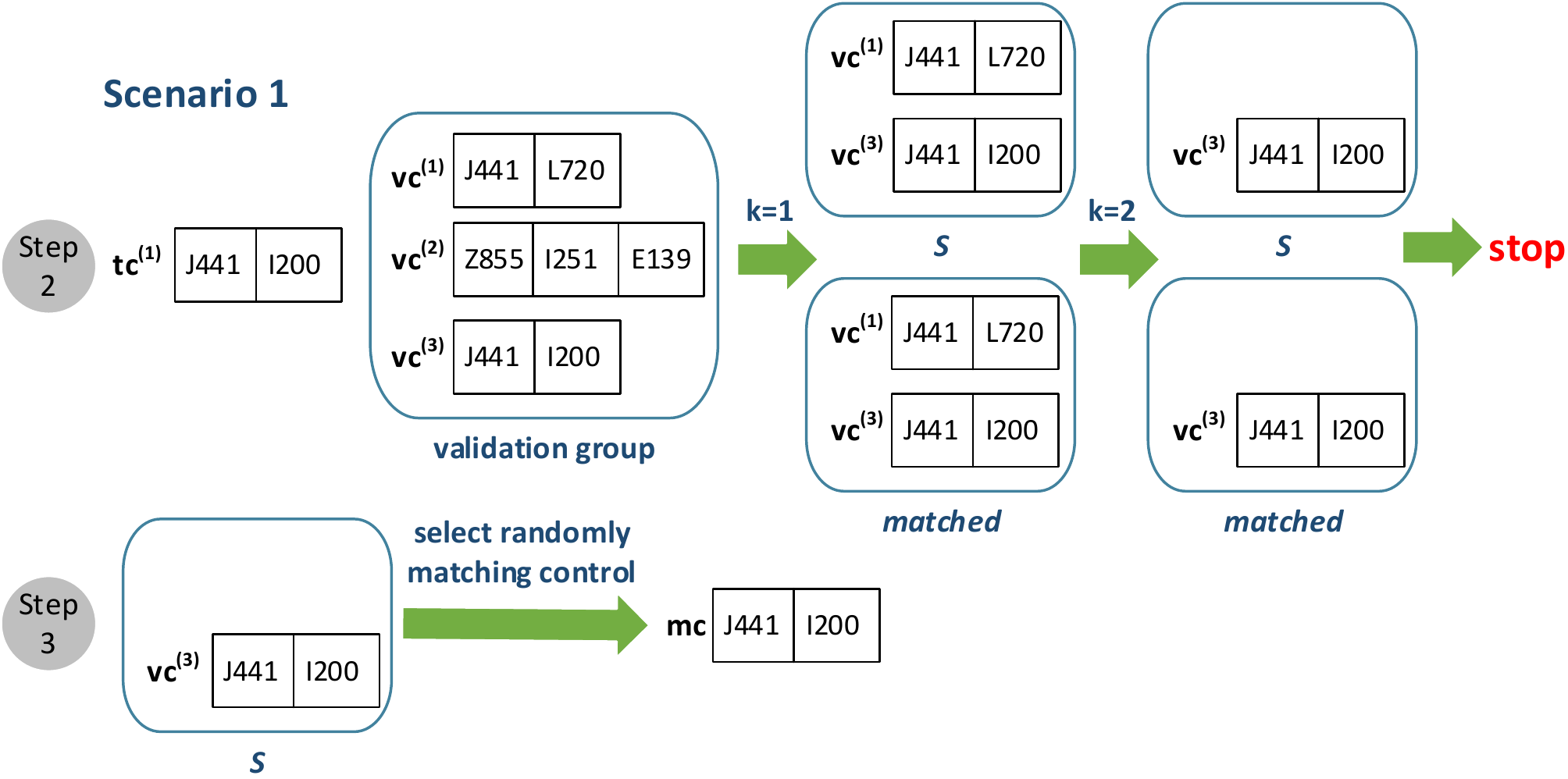}
\par\end{centering}
\caption{\label{fig:SM-Case-1}We can find the controls that match all ICD
codes in $tc^{\left(1\right)}$: $vc^{\left(3\right)}$ is identical
to $tc^{\left(1\right)}$. Note that $tc^{\left(1\right)}$ is a case
and $vc^{\left(3\right)}$ is a control}
\end{figure}

\begin{figure}
\begin{centering}
\includegraphics[scale=0.4]{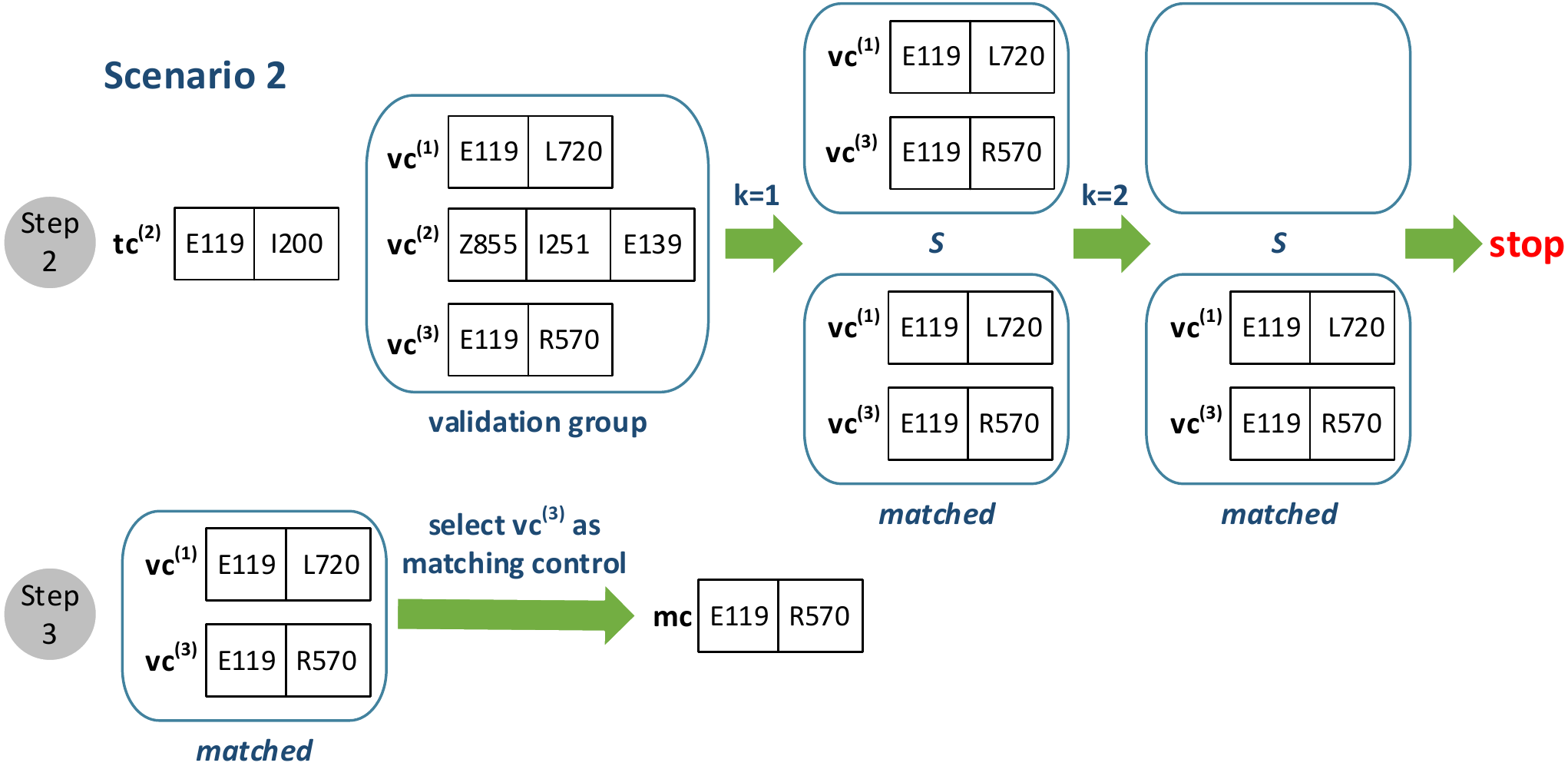}
\par\end{centering}
\caption{\label{fig:WVM-Case-2}We can find the controls that match at least
one ICD code in $tc^{\left(2\right)}$. SM selects $vc^{\left(3\right)}$
as the matching control because the ICD code vector of R570 is close
to the ICD code vector of I200}

\end{figure}

\begin{figure}
\begin{centering}
\includegraphics[scale=0.4]{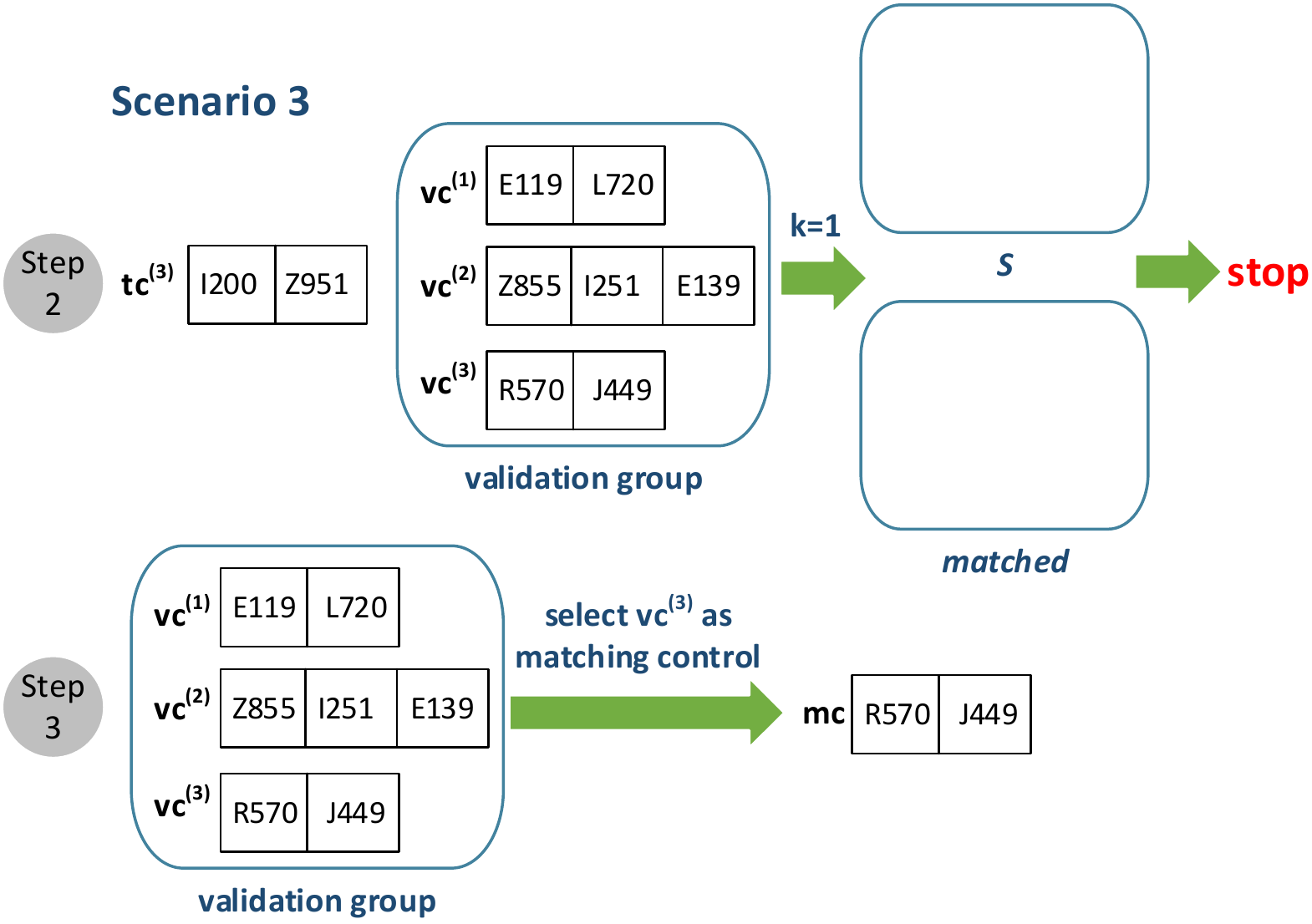}
\par\end{centering}
\caption{\label{fig:WVM-Case-3}We cannot find any control that matches the
first ICD code in $tc^{\left(3\right)}$. SM selects $vc^{\left(3\right)}$
as the matching control because I200 and R570 are similar}

\end{figure}

We provide three examples as shown in Figures \ref{fig:SM-Case-1}-\ref{fig:WVM-Case-3}
to demonstrate the three scenarios processed by steps 2 and 3.\vspace{-0.2cm}

\vspace{-0.2cm}

\section{Experiments\vspace{-0.2cm}
}

\subsection{Data\vspace{-0.2cm}
}

The dataset is a cancer cohort of more than 220,000 patients (58.2\%
males, median age 71), collected between 1997 and 2012 from a state-wide
cancer registry including 21 hospitals in Australia. The data attributes
include patient demographic and diagnoses indicated by ICD-10 codes.
There is no information of labs, procedures, and drugs. A subset of
data for control matching was selected as follows: 1) we removed the
ICD codes occurring less than 30 times; 2) we only kept the admissions
that have at least one ICD code; and 3) we removed the admissions
that have discharge date after 2008/12/31. The remaining dataset contains
1,810,967 admissions (216,844 unique patients).\vspace{-0.2cm}

\subsection{Construction of case group and control group\label{subsec:Construction-of-case}\vspace{-0.2cm}
}

First, we selected randomly a combination of a hospital $H$ and an
admission year $Y$ from the dataset. We then selected the admissions
that have the same hospital and admission year as $H$ and $Y$. Note
that each admission is associated with a different patient. This set
is called $HY$. Final, we constructed the case group by sampling
200 admissions from $HY$. The remaining admissions in $HY$ were
used to construct the control group. The admissions in the control
group are associated with the patients who are different from those
in the case group.\vspace{-0.2cm}

\subsection{Performance evaluation\vspace{-0.2cm}
}

We ran four methods randomly in 150 times. To have a fair comparison,
all baselines were performed after step 1 in SM was done (i.e., after
we obtained a validation group). Each time of running, we obtained
200 cases and 200 matching controls and measured the agreement of
the two cohorts on the two clinical outcomes.\vspace{-0.2cm}

\subsubsection{Readmission matching accuracy\vspace{-0.2cm}
}

Let $trial^{\left(i\right)}=\left(y_{1},y_{2},...,y_{200}\right)$
be a set of readmission statuses of 200 cases at the iteration $i$
(true values). $y_{r}=trial_{r}^{\left(i\right)}$ ($r\in\left[1,200\right]$)
is one of four readmission statuses: ``Missing'' (0.01\%), ``Readmitted
within 28 days to another facility'' (4.07\%), ``Readmitted within
28 days to the same facility'' (23.03\%), and ``Not formally readmitted
within 28 days'' (72.89\%). Let $match^{\left(i\right)}$ be a set
of readmission statuses of 200 matching controls at the iteration
$i$ (predicted values). The readmission matching accuracy at the
iteration $i$ is computed as follows.\vspace{-0.2cm}

\begin{equation}
acc^{\left(i\right)}={\displaystyle \sum_{r=1}^{200}\omega\left(trial_{r}^{\left(i\right)},match_{r}^{\left(i\right)}\right)/200,}\label{eq:match_acc}
\end{equation}
\vspace{-0.2cm}

where $\omega\left(trial_{r}^{\left(i\right)},match_{r}^{\left(i\right)}\right)=\begin{cases}
0 & \text{if \ensuremath{trial_{r}^{\left(i\right)}\neq match_{r}^{\left(i\right)}}}\\
1 & \text{if \ensuremath{trial_{r}^{\left(i\right)}=match_{r}^{\left(i\right)}}}
\end{cases}$.

Table \ref{tab:Readmission-matching-accuracy} reports the average
readmission matching accuracy for 150 running times of each method.
Our proposed method (WVM) is better than non-embedding ICD code methods
(PCM and HDM). Although both CSM and WVM learn ICD code vectors, CSM\footnote{For a given case $tc^{\left(i\right)}$, CSM sums up the vectors of
all ICD codes in $tc^{\left(i\right)}$ into a single vector. For
each control, it performs the same task. It then determines the similarity
between $tc^{\left(i\right)}$ and a control by computing the cosine
distance between their summed vectors.} however does not consider the importance of the orderliness of ICD
codes within a code sequence; its accuracy is thus lower than WVM.\vspace{-0.2cm}

\subsubsection{Incidence rate (IR) error for cancer mortality\vspace{-0.2cm}
}

Let $trial^{\left(i\right)}=\left(\left(s_{1},t_{1}\right),\left(s_{2},t_{2}\right),...,\left(s_{200},t_{200}\right)\right)$
be a set of 2-tuples (\textit{discharge date}, \textit{death date})
of 200 cases at the iteration $i$. Each case has a discharge date
and a death date that may be \textit{null}. For example, the first
case in the case group has (\textit{discharge date}, \textit{death
date}) of (2005/07/11, 2008/05/12); the second case has (\textit{discharge
date}, \textit{death date}) of (2005/10/27, \textit{null}). The incidence
rate of the case group at the iteration $i$ (true value) is computed
as follows.\vspace{-0.2cm}

\begin{equation}
IR\left(trial^{\left(i\right)}\right)=\frac{count\left(\text{\# of death cases}\right)}{{\displaystyle \sum_{r=1,t_{r}\neq null}^{200}\left(t_{r}-s_{r}\right)+\sum_{r=1,t_{r}=null}^{200}}\left(d_{censor}-s_{r}\right)},\label{eq:ir}
\end{equation}
\vspace{-0.2cm}

where $count\left(\text{\# of death cases}\right)$ is the number
of cases that have death date (i.e., their death dates are not \textit{null}),
$t_{r}$ and $s_{r}$ are death date and discharge date respectively,
and $d_{censor}$ is the censoring date (i.e., the end date of our
study, that is 2008/12/31).

Similarly, we can compute the incidence rate of 200 matching controls
at the iteration $i$ (predicted value), called $IR\left(match^{\left(i\right)}\right)$.
The incidence rate error (absolute error) at the iteration $i$ is
computed as follows.\vspace{-0.2cm}

\begin{equation}
IR_{err}^{\left(i\right)}=\left|IR\left(trial^{\left(i\right)}\right)-IR\left(match^{\left(i\right)}\right)\right|\label{eq:ir_err}
\end{equation}
\vspace{-0.2cm}

Table \ref{tab:Incidence-rate-error} reports the mean incidence rate
error of each method in 150 running times. Again, the mean incidence
rate error of our proposed method (WVM) has the smallest value. \vspace{-0.2cm}

\begin{figure}[th]
\begin{minipage}[t]{0.49\linewidth}%
\captionof{table}{Readmission matching accuracy}\label{tab:Readmission-matching-accuracy}
\begin{center}
\begin{tabular}{|l|c|}
\hline 
\textbf{Method} & \textbf{Accuracy}\tabularnewline
\hline 
\hline 
PCM \cite{lee2015personalized} & 0.7565$\pm$0.0070\tabularnewline
\hline 
HDM \cite{hielscher2014using} & 0.7693$\pm$0.0067\tabularnewline
\hline 
CSM \cite{choi2016medical} & 0.7755$\pm$0.0067\tabularnewline
\hline 
WVM & \textbf{0.7952}$\pm$0.0068\tabularnewline
\hline 
\end{tabular}
\par\end{center}%
\end{minipage}%
\begin{minipage}[t]{0.49\linewidth}%
\captionof{table}{Incidence rate error}\label{tab:Incidence-rate-error}
\begin{center}
\begin{tabular}{|l|c|}
\hline 
\textbf{Method} & \textbf{IR Error}\tabularnewline
\hline 
\hline 
PCM \cite{lee2015personalized} & 0.0342$\pm$0.0032\tabularnewline
\hline 
HDM \cite{hielscher2014using} & 0.0316$\pm$0.0026\tabularnewline
\hline 
CSM \cite{choi2016medical} & 0.0322$\pm$0.0028\tabularnewline
\hline 
WVM & \textbf{0.0299}$\pm$0.0022\tabularnewline
\hline 
\end{tabular}
\par\end{center}%
\end{minipage}

\end{figure}
\vspace{-0.2cm}

\section{Conclusion\vspace{-0.2cm}
}

We have introduced WVM, a case-control matching method that leverages
both representational similarity among ICD-10 codes and the sequential
structure of coding in each admission. The evaluation on two similarity
measures based on clinical outcomes, namely readmission matching accuracy
and incidence rate error for cancer mortality, proves that WVM constitutes
an effective solution for patient similarity matching in a large cancer
cohort. In practice, it means that WVM can identify a control cohort
better matching the case cohort, hence minimizing the potential bias
between the two cohorts. This enables more effective experiment or
quasi-experiment designs using a large coded dataset that is similar
to ours.

\bibliographystyle{plain}
\bibliography{nips_ml4hc_2016}

\end{document}